\documentclass[runningheads]{llncs}
\usepackage{graphicx}

\usepackage{bm}
\usepackage{booktabs}
\usepackage{xspace}
\usepackage{float}

\usepackage{tikz}
\usepackage{comment}
\usepackage{amsmath,amssymb} 
\usepackage{color}
\usepackage[capitalize]{cleveref}
\crefname{section}{Sec.}{Secs.}
\Crefname{section}{Section}{Sections}
\Crefname{table}{Table}{Tables}
\crefname{table}{Tab.}{Tabs.}

\usepackage[labelsep=period]{caption}
\RequirePackage{enumitem}
\setlist[itemize]{nosep}

\usepackage[accsupp]{axessibility}  

\begin{document}
\pagestyle{headings}
\mainmatter
\def\ECCVSubNumber{2747}  

\title{Towards High-Fidelity Single-view Holistic Reconstruction of Indoor Scenes}

\titlerunning{InstPIFu} 

\author{Haolin Liu\inst{1,2}\thanks{Both authors contributed equally to this paper.} \and
Yujian Zheng\inst{1,2\star} \and Guanying Chen\inst{1,2} \and Shuguang Cui\inst{1,2} \and
Xiaoguang Han\inst{1,2}\thanks{\textbf{Corresponding Email}: hanxiaoguang@cuhk.edu.cn} }
\authorrunning{Liu et al.}
%
\institute{School of Science and Engineering, CUHK-Shenzhen \and
The Future Network of Intelligence Institute, CUHK-Shenzhen\\
}
\maketitle

\newcommand{\gy}[1]{\textcolor{red}{{[GY: #1]}}}
\newcommand{\yj}[1]{\textcolor{red}{{[YJ: #1]}}}
\newcommand{\todo}[1]{\textcolor{red}{{[todo: #1]}}}

\newcommand{\methodName}{InstPIFu\xspace}
\newcommand{\MethodNameLong}{Instance-aligned Implicit Function\xspace}
\newcommand{\methodnamelong}{instance-aligned implicit function\xspace}
\newcommand{\instpixelaligned}{instance-aligned\xspace}
\newcommand{\Instpixelaligned}{Instance-aligned\xspace}
\newcommand{\pixelaligned}{pixel-aligned\xspace}
\newcommand{\Pixelaligned}{Pixel-aligned\xspace}
\newcommand{\instattention}{instance-aligned attention\xspace}
\newcommand{\InstAttention}{Instance-aligned Attention\xspace}
\newcommand{\attentionmodule}{\instattention module\xspace}
\newcommand{\AttentionModule}{\InstAttention Module\xspace}
\newcommand{\spatialsupervision}{spatial-guided supervision\xspace}
\newcommand{\Spatialsupervision}{Spatial-guided supervision\xspace}
\newcommand{\pixelwise}{pixel-wise\xspace}
\newcommand{\channelwise}{channel-wise\xspace}
\newcommand{\sunrgbd}{SUN RGB-D\xspace}
\newcommand{\pixthreeD}{Pix3D\xspace}
\newcommand{\threeDfront}{3D-FRONT\xspace}
\newcommand{\threeDfuture}{3D-FUTURE\xspace}

\newcommand{\ie}{\textit{i}.\textit{e}.}
\newcommand{\eg}{\textit{e}.\textit{g}.}

\newcommand{\sota}{state-of-the-art\xspace}

\setlength{\textfloatsep}{5pt}
\begin{abstract}
   We present a new framework to reconstruct holistic 3D indoor scenes including both room background and indoor objects from single-view images. 
   Existing methods can only produce 3D shapes of indoor objects with limited geometry quality because of the heavy occlusion of indoor scenes. 
   To solve this, we propose an \instpixelaligned implicit function (\methodName) for detailed object reconstruction.
   Combining with instance-aligned attention module, our method is empowered to decouple mixed local features toward the occluded instances. 
   Additionally, unlike previous methods that simply represents the room background as a 3D bounding box, depth map or a set of planes, we recover the fine geometry of the background via implicit representation. 
   Extensive experiments on the \sunrgbd, \pixthreeD, \threeDfuture, and \threeDfront datasets demonstrate that our method outperforms existing approaches in both background and foreground object reconstruction.
   Our code and model will be made publicly available.
\end{abstract}

\begin{figure}[htbp]
	\includegraphics[width=1\linewidth]{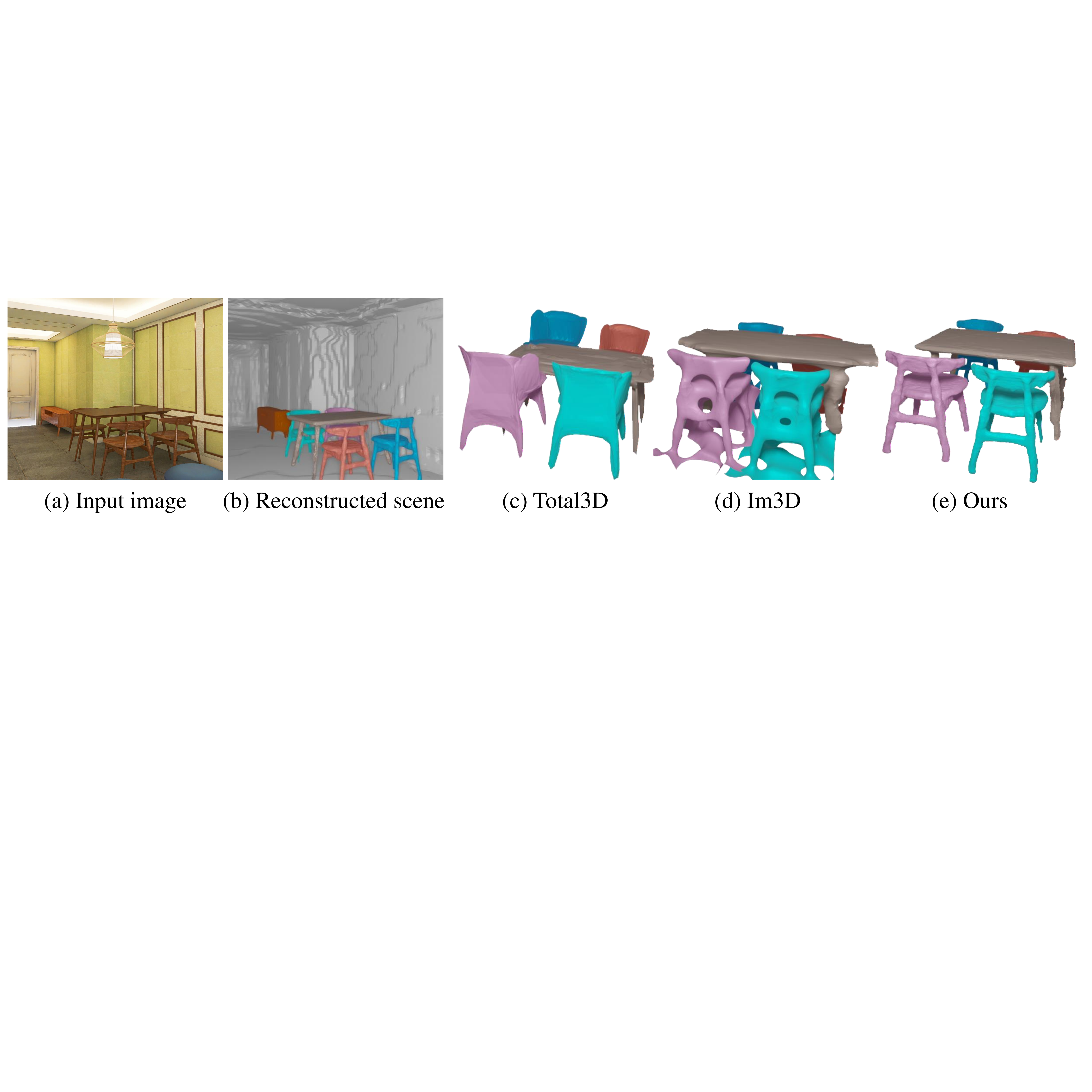}
	\caption{Given a single indoor scene image, we reconstruct the holistic scene with detailed geometry, including the room background and indoor objects. From left to right: input image, the scene reconstructed by our method, results of Total3D~\cite{total3d}, Im3D~\cite{im3d} and our method in a different camera pose.}
	\label{fig:teaser}
\end{figure}
\section{Introduction}
\label{sec:intro}

With the development of virtual reality (VR) and augmented reality (AR), the requirements for understanding and digitizing real-world 3D scenes are getting higher, especially for the indoor environment. 
If reconstructing the holistic indoor scene can be as simple as taking a picture using a mobile phone, we can efficiently generate a large scale of high-quality 3D content and further promote the development of VR and AR.
Also, robots can better understand the real-world with the advance of single-view scene reconstruction.
Hence, the problem of holistic indoor scene reconstruction from a single image has attracted considerable attention in recent years.

Early methods simplify this problem as estimating the room layout~\cite{hedau2009recovering,lee2009geometric,mallya2015learning,dasgupta2016delay,ren2016coarse} and indoor objects~\cite{du2018learning,huang2018cooperative,chen2019holistic++} as 3D bounding boxes. 
However, such a coarse representation can only provide scene context information but cannot provide shape-level reconstruction.
Mesh retrieval based approaches~\cite{izadinia2017im2cad,hueting2017seethrough,huang2018holistic} improve the object shapes by substituting the 3D object boxes with meshes searched from a database. Due to the various categories and appearances of indoor objects, the size and diversity of the database directly influence the accuracy and time efficiency of these methods.
 
Inspired by learning-based shape reconstruction methods, voxel representation~\cite{li2019silhouette,tulsiani2018factoring,kulkarni20193d} is first applied to recover the 3D geometry of indoor scenes, but the shape quality is far from satisfactory due to the limited resolution.
Mesh R-CNN~\cite{gkioxari2019mesh} can reconstruct meshes for multiple instances from a single-view image, but lacks of scene understanding.
Recently, Total3D\cite{total3d} and Im3D\cite{im3d} are proposed to reconstruct the 3D indoor scene from a single image, where the instance-level objects are represented in the form of explicit mesh and implicit surface, respectively. 
Although they have achieved \sota results on this task, they still have the following limitation.
First, they often output shapes lacking details,
due to the issue of limited training data and the use of global image feature for shape reconstruction. 
Second, the room layout in their methods is expressed as a simplified representation (\ie, the 3D bounding box) which cannot recover backgrounds with complex geometries, like non-planar surfaces. 

Recently, pixel-aligned implicit function (PIFu) has achieved promising results for detailed and generalizable 3D human reconstruction from a single image~\cite{saito2019pifu}.
Motivated by the success of PIFu, we address the limitations of existing methods by introducing an \emph{\methodnamelong} (\methodName) for holistic and detailed indoor scene reconstruction from a single image. 
Note that pixel-aligned feature cannot be straightforwardly applied to indoor scene reconstruction, as objects (\eg, sofa, chair, bed, and other furniture) are often occluded in a cluttered scene (see~\cref{fig:teaser}), such that the extracted local feature might contain mixed information of multiple objects. 
It is sub-optimal to directly use such a contaminated local feature for implicit surface reconstruction.
To tackle this problem, we introduce an \emph{\attentionmodule}, consisting of \emph{attentional channel filtering}, and \emph{\spatialsupervision} strategies, to decouple the mixed local features for different instances in the overlapping regions.

Unlike previous methods that simply recover the room layout as a 3D bounding box~\cite{hedau2009recovering,lee2009geometric,mallya2015learning,dasgupta2016delay,ren2016coarse,total3d,im3d}, sparse depth~\cite{tulsiani2018factoring} or room layout structure~\cite{zou2018layoutnet,stekovic2020general,yang2022learning} without  non-planar geometry, our implicit surface representation allows the detailed shape reconstruction of the room background (\eg, floor, wall, and ceiling).
Compared with existing approaches that encode the latent shape code with global image features~\cite{total3d,im3d}, the instance-aligned local features utilized in our encoder help alleviate the over-fitting problem and recover more detailed geometry of indoor objects. Extensive experiments on the SUN RGB-D, Pix3D, 3D-FUTURE, and 3D-FRONT datasets demonstrate the superiority of our method.

The key contributions of this paper are summarized as follows:

\begin{itemize}[leftmargin=*]
	\item We introduce a new pipeline to reconstruct the holistic and detailed 3D indoor scene from a single RGB image using implicit representation. To our best knowledge, this is the first system that uses pixel-aligned feature to recover the 3D indoor scene from a single view.

	\item We are the first to attempt to reconstruct the room background via implicit representation. Compared to previous methods that represent room layout as a 3D box, depth map or a set of planes, our method is capable to recover background with more complex geometries, like non-planar surfaces.
	
	\item We propose a new method, called \methodName,  to use the \instpixelaligned feature, extracted by a novel \attentionmodule, for detailed indoor object reconstruction. 
	Our method is more robust to object occlusion and has a better generalization ability on real-world datasets.
	
	\item Our method achieves \sota performance on both the synthetic and real-world indoor scene datasets. 
\end{itemize}

\section{Related Work}
\label{sec:related}

\paragraph{Single-view indoor scene reconstruction} 
The long-standing problem of indoor scene reconstruction from a single image aims to construct the holistic 3D scene, which entails room layout estimation, object detection and pose estimation, as well as 3D shape reconstruction. 
Early works first recover the room layout as a 3D room bounding box~\cite{hedau2009recovering,lee2009geometric,mallya2015learning,dasgupta2016delay,ren2016coarse}. 
Follow-up works make rapid progress toward object pose recovery~\cite{du2018learning,huang2018cooperative,chen2019holistic++}, but still represent objects as 3D boxes without shape details. 

To recover object shapes, some methods search for models with a similar appearance from a database~\cite{izadinia2017im2cad,hueting2017seethrough,huang2018holistic}. 
However, the mismatch between objects in images and the database often leads to unsatisfactory results. 
Other methods~\cite{li2019silhouette,tulsiani2018factoring,kulkarni20193d} try to reconstruct the voxel representation for each object instance, but they are subjected to the problem of limited resolution. 
Mesh R-CNN~\cite{gkioxari2019mesh} is capable to reconstruct meshes for multiple objects from a single-view image, but ignores scene understanding.
To overcome the above limitations of previous solutions, Total3D~\cite{total3d} proposes an end-to-end system to jointly reconstruct room layout, object bounding boxes, and meshes from a single image. 
But its mesh generation network can only produce non-watertight mesh when handling shapes with complex topology. 
The following Im3D~\cite{im3d} represents each object with the implicit surface function that can be converted to a watertight mesh via marching cube algorithm while preserving geometry details in the meantime. 
However, the state-of-the-art solution of Im3D~\cite{im3d} still suffers from shape over-fitting due to the problem of limited training data and the use of global image features for shape reconstruction.

\paragraph{Room background representation} 
Early methods~\cite{hedau2009recovering,lee2009geometric,mallya2015learning,dasgupta2016delay,ren2016coarse} simply recover the room background as a 3D bounding box, but room is usually not a cuboid. The state-of-the-art single-view indoor scene reconstruction methods~\cite{total3d,im3d} are still using this representation for room background. 
~\cite{tulsiani2018factoring} predicts the background via depth estimation, which recovers more details for background. However, the accuracy of background depth estimation is far from satisfactory because of the occlusion of foreground, \ie, indoor objects.
Recent works try to reconstruct the room layout structure~\cite{zou2018layoutnet,stekovic2020general,yang2022learning} with the assumption that the background of the room (\eg, floor, wall, and ceiling) is mainly composed of planes. Hence, only planar geometry can be recovered and nonplanar information is missed by these methods.

\paragraph{Learning-based 3D shape reconstruction} 
Recent learning-based methods have adopted different surface representations for 3D shape reconstruction, such as voxel, mesh, point cloud, patches, primitives, and implicit surface. 

Voxel-based methods~\cite{3D-R2N2,LiaoDG18,Wallace,riegler2017octnet,tatarchenko2017octree,wang2018adaptive} benefit from 2D CNNs because of the regularity of the voxel representation, but suffer from the balance between resolution and efficiency. 
Mesh-based methods reconstruct the mesh of an object through deforming a template (\eg, a unit sphere), but the topology of the obtained mesh is restricted~\cite{wang2018pixel2mesh,groueix2018,Junyi,kato2018neural}. 
To modify the topology, some approaches learn to remove extra edges and vertices~\cite{Junyi,tang2019skeleton,total3d}, which results in non-watertight meshes. 
Methods based on point cloud~\cite{fan2017point,mandikal20183d,kurenkov2018deformnet,navaneet2019capnet}, patches~\cite{groueix2018,wang2018adaptive}, and primitives~\cite{tian2019learning,tulsiani2017learning,paschalidou2019superquadrics,deprelle2019learning} are adaptable to complex topology, but require post-processing to convert to structural representations. However, the post-processing is difficult to preserve the detailed geometry of the shape. 
Recently, implicit surface function ~\cite{park2019deepsdf,chen2019learning,michalkiewicz2019deep,xu2019disn,mescheder2019occupancy} has been widely adopted as it can achieve detailed reconstruction for shape with an arbitrary topology and is easy to be converted to fine mesh.

\paragraph{Pixel-aligned image features} 
Single-view implicit surface reconstruction methods often adopt an encoder-decoder pipeline and learn a latent code from the input image for shape recovery. 
For time and memory efficiency, global image feature~\cite{park2019deepsdf,chen2019learning,niemeyer2020differentiable,mescheder2019occupancy,dupont2020equivariant} is often adopted, but it cannot recover the local detailed information existed in the input image. 
As a result, coarse results often occur in these approaches. 
Recently, pixel-aligned local image features have been demonstrated to recover complex geometries from a single view~\cite{saito2019pifu,xu2019disn}.

\begin{figure*}[t]
	\centering
	\includegraphics[width=1.0\textwidth]  
	{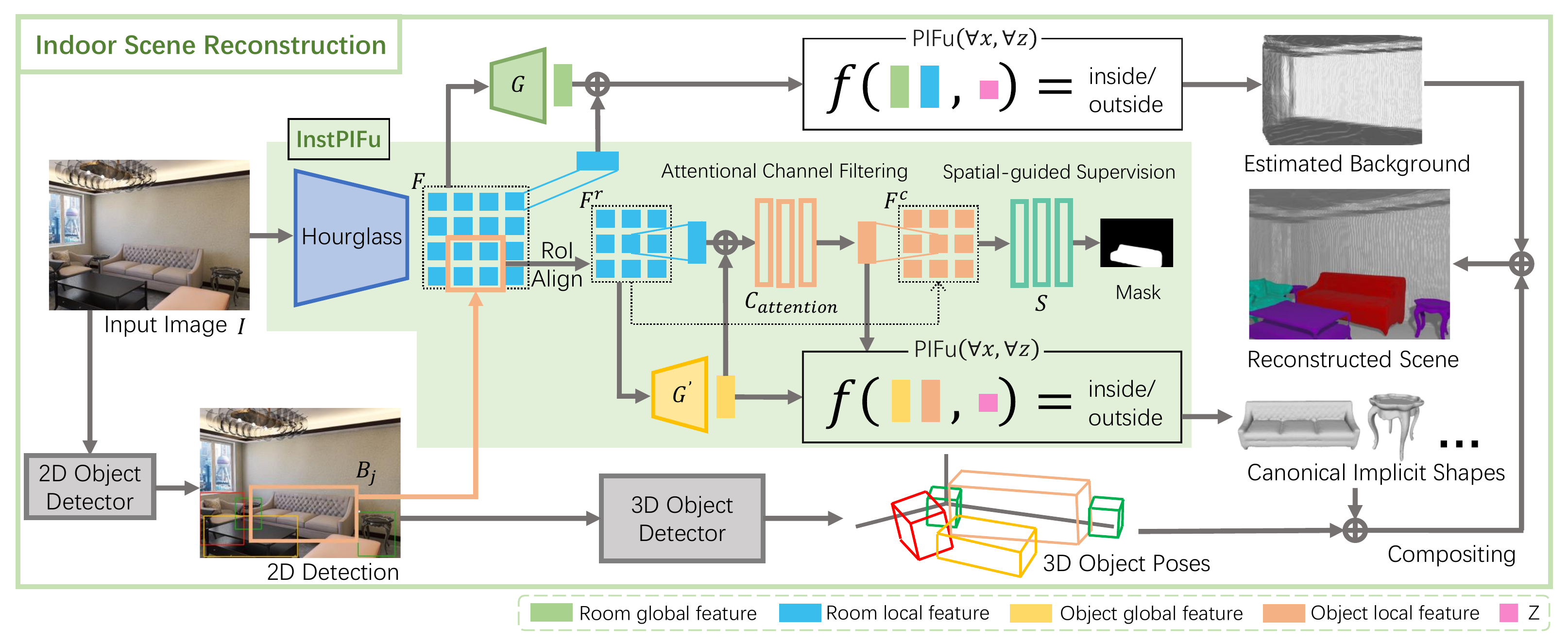}
	\caption{Overview of the proposed \methodName. Given a single indoor scene image as input, our method simultaneously performs room background estimation, object detection and camera pose estimation, as well as detailed 3D object reconstruction.}
	\label{fig:overview}
\end{figure*}

\section{\Instpixelaligned Implicit Representation}
\label{sec:instpifu}
In this section, we first review the \pixelaligned implicit function (PIFu) and point out its limitation in dealing with the occluded object in the indoor scene. We then introduce our \methodnamelong to perform better indoor object reconstruction where objects are often occluded in the cluttered scene.

\subsection{Review of \Pixelaligned Implicit Modeling}
Single-view scene reconstruction benefits from implicit representation~\cite{im3d}, but the usage of global image features often causes coarse results. 
PIFu with pixel-aligned local features has been witnessed to recover detailed shapes in 3D human reconstruction~\cite{saito2019pifu}. 

A 3D surface can be defined by an implicit function as a level set of function $f$, \eg $f(X)=0$, where $X$ is a 3D point. 
Similarly, a pixel-aligned implicit function $f$, represented by multi-layer perceptrons (MLPs), defines the surface as a level set of
\begin{equation}
\label{eqn:00}
\begin{aligned}
f(F(x), z(X))=s: s \in \mathbb{R},
\end{aligned}
\end{equation}
where $x=\pi(X)$ gives the 2D image projection point of $X$, $F(x)=g(I(x))$ is the local image feature at $x$ extracted by a fully convolutional image encoder $g$, and $z(X)$ is the depth value in weak-perspective camera coordinate. 
We observe that adding the global image feature as an extra input helps in shape reconstruction. The adapted PIFu used in this work is defined as
\begin{equation}
\label{eqn:01}
\begin{aligned}
f(F(x), F^G(I), z(X))=s: s \in \mathbb{R},
\end{aligned}
\end{equation}
where $F^G(I)$ represents the global features of image $I$ encoded by $G$.

\subsection{Limitation of \Pixelaligned Feature}
Although PIFu demonstrates detailed reconstruction results in single human reconstruction, applying PIFu for indoor object reconstruction straightforwardly is not good, as it suffers a lot from the object occlusion that leads to feature ambiguity. 
Multiple 3D points belonging to different objects can be projected into similar 2D image location and get the same local image feature, such that the local feature will contain mixed information from different instances, which is not desirable for shape reconstruction.

\begin{figure}[t]
    \centering
	\includegraphics[width=1.0\linewidth]{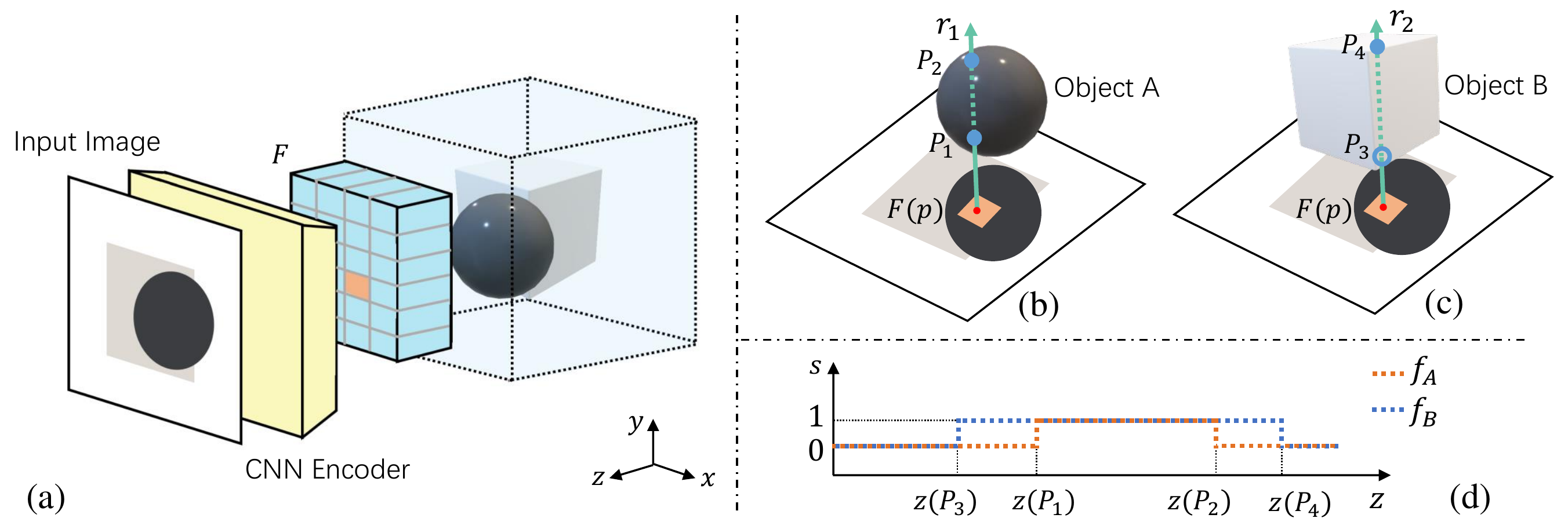}
	\caption{Occlusion causes local feature ambiguity among different objects. (a) A scene contains two objects, and $F$ is the extracted local feature from the image. (b)-(c) Object reconstruction in canonical coordinate system, where points along the rays are projected at $p$ to sample local feature $F(p)$. (d) Variations of occupancy $s$ with depth $z$ along the ray $r_1$ and $r_2$ for $f_A$ and $f_B$.}
	\label{fig:occlusion}
\end{figure}

As an example in \cref{fig:occlusion}~(a), a scene consists of a sphere A and and a cube B, where A occludes B in the captured image.
\cref{fig:occlusion} (b)-(c) show that when sampling pixel-aligned features for 3D points, points along the ray $r_1$ and $r_2$ (\eg, $P$) are all projected at the point $p$ in the overlapping region. 
This means that the same local feature $F(p)$ will be used to compute the occupancy value $s$ for implicit function of A ($f_A$) and B ($f_B$), \ie, $s=f_{A}(F(p),z(P))$ and $s=f_{B}(F(p),z(P))$. 
As PIFu implemented $f_A$ and $f_B$ using the same MLPs, adopting the same local feature $F(p)$ raises feature ambiguity in occupancy estimation for A and B. 
This is illustrated in \cref{fig:occlusion}~(d), where variations of $s$ with $z$ for $f_A$ and $f_B$ are apparently different. 
Note that here we simply represent the PIFu $f$ as an ideal occupancy field where levels of points inside the object are $1$, otherwise $0$. 

One possible solution might be adding the global feature of the instance as extra inputs to the shape decoder.
But only using global features to tackle the ambiguity in occlusion region is not enough (see our ablation study). Because the local features still contain mixed information from different instances.

\subsection{\Instpixelaligned Feature Concentration}
To address the above limitation of PIFu, we propose \methodName, which adopts an \instattention module to disentangle the mixed feature information caused by object occlusion, for indoor object reconstruction.
The proposed \attentionmodule reduces the ambiguity of the local image feature by three sequential steps, \ie, \emph{RoI alignment}, \emph{attentional channel filtering}, and \emph{\spatialsupervision} (see~\cref{fig:overview}).

\newcommand{\instancesubscript}{j}
\newcommand{\roialign}{\text{RoIAlign}}
\newcommand{\imagefeature}{F}
\newcommand{\roifeature}{F^{r}}
\newcommand{\bbox}{B}
\newcommand{\pixelwiseattention}{P_\text{{attention}}}
\newcommand{\instanceimageencoder}{G'}
\newcommand{\image}{I}
\newcommand{\globalinstancefeature}{\instanceimageencoder(\roifeature)}
\newcommand{\globalinstancefeatureabbr}{\globalinstancefeature}
\newcommand{\pixelwisefeature}{F^p}
\newcommand{\channelwiseattention}{C_\text{attention}}
\newcommand{\channelwisefeature}{F^c}
\newcommand{\instalignfunction}{f_{o}}
\newcommand{\spatialguide}{S}
\newcommand{\mask}{M}

\paragraph{RoI alignment} 
The first step is to extract instance-related features for each instance.
A straight-forward solution is to extract features independently from the cropped image patch of each target instance. However, it is inefficient when there are multiple objects in a cluttered scene, and the useful scene contextual information will be ignored. 
Instead, we follow Mask R-CNN \cite{he2017mask} to use RoI alignment for instance-related feature extraction.
Given an image $\image$ and the 2D bounding box $\bbox_\instancesubscript$ of an instance $\instancesubscript$, we first crop out the corresponding local features of the region of interest (RoI) from the whole pixel-aligned feature map $\imagefeature$ and align them to $\roifeature$ as~\cite{he2017mask}
\begin{equation}
\begin{aligned}
\roifeature = \roialign(\imagefeature, \bbox_\instancesubscript).
\end{aligned}
\end{equation}
Note that $\roifeature$ has a fixed size of $W_r \times H_r$ for input feature maps with different shapes and the 2D bounding box $B_j$ of object $j$ is obtained by a Faster R-CNN detector~\cite{ren2015faster}. We then extract a global instance feature for instance $\instancesubscript$ as $\globalinstancefeature$, where $\instanceimageencoder$ is a global instance image encoder.
The global instance feature will be used to compute the channel-wise attention for local feature filtering.

\paragraph{Attentional channel filtering} 
Each local feature in the aligned RoI feature map $\roifeature$ will be concatenated with the global instance feature $\globalinstancefeatureabbr$ as input for a channel-wise attention layer, similar to the Squeeze-and-Excitation block in~\cite{hu2018squeeze} structurally, to generate an attention map with the same channel number of $L_c$ as the local feature.
This attention map will multiply with the local feature to filter out irrelevant feature by channel filtering to allow the updated local feature to concentrate on the target instance.
This operation can be expressed as 
\begin{equation}
\begin{aligned}
\channelwisefeature(x) = \channelwiseattention(\roifeature(x), \globalinstancefeatureabbr) \times \roifeature(x),
\end{aligned}
\end{equation}
where $\channelwisefeature(x)$ is the filtered local image feature at the 2D projection $x$ of a 3D point $X$ for instance $\instancesubscript$. Note that $F^r(x)$ in Eq. (4) adopts bilinear interpolation to access the features, and $x$ in $F^r(x)$ should be shifted and scaled as well.

\paragraph{\Spatialsupervision} 
To better guide the learning of the channel filtering, we need a module that can encourage the filtered feature to focus more on the target instance. Thus, we exploit a \spatialsupervision on the the filtered local feature map $\channelwisefeature$
that is the output of the channel-wise attention layer with the same shape as $\roifeature$.
The Feature map $\channelwisefeature$ will be fed into a fully convolutional layer $\spatialguide$ to estimate a complete mask $\mask$ for the target instance, \ie, $\mask = \spatialguide(\channelwisefeature)$. This \spatialsupervision can filter out irrelevant information out of the mask.

\paragraph{\Instpixelaligned implicit function}
Given the \instpixelaligned feature, we define \methodName $\instalignfunction$ as 
\begin{equation}
\label{eqn:02}
\begin{aligned}
\instalignfunction(\channelwisefeature(x), \globalinstancefeatureabbr, z(X))=s: s \in \mathbb{R}.
\end{aligned}
\end{equation}
By applying the proposed \attentionmodule for decoupling the mixed local feature,
compared with PIFu, the local feature used in our \methodName provides more discriminative information for accurate and detailed shape reconstruction. And this can be demonstrated by our ablation study. 

\section{Holistic Indoor Scene Reconstruction}
Given a single image of an indoor scene, we aim to recover the holistic and detailed 3D scene in implicit representation (see~\cref{fig:overview}).
This problem is normally divided into several sub-tasks, including room background estimation, 3D object detection (pose estimation), as well as instance-level object reconstruction~\cite{total3d,im3d}. 
We first process these three tasks individually and then perform scene compositing for holistic scene reconstruction. 
Note that our method recovers the room background with geometry details instead of just a simplified 3D bounding box. 

\subsection{Room Background Estimation} 
Room is usually not a cuboid. Thus, it is inappropriate to represent the room background as a 3D bounding box like~\cite{total3d,im3d}. Depth map~\cite{tulsiani2018factoring} is also not an ideal representation, because the accuracy of background depth estimation is heavily influenced by the occlusions of indoor objects in front of the background. Also, methods~\cite{stekovic2020general,yang2022learning} based on plane detection cannot recover small planes and non-planar background geometries. To address the above issues, we explore to use the implicit representation for room background reconstruction in this work. 

The ground-truth room surface is represented as a $0.5$ level set and then discretized to a 3D occupancy field:
\begin{equation}
\label{eq:level_set}
\begin{aligned}
f_{r}^{*}(X)= \begin{cases}1, & \text { if } X \text { is inside the room } \\ 0, & \text { otherwise }\end{cases}.
\end{aligned}
\end{equation}
Compared with indoor objects that have various styles and complicated geometries, the shape of the room background is much simpler. 
We find that applying the adapted PIFu (see~\cref{eqn:01}) which takes pixel-aligned features and global features for room background reconstruction already achieves good results. 
We train our room estimation PIFu $f_{r}$ by minimizing the average of mean squared error (MSE):
\begin{equation}
\begin{aligned}
\mathcal{L}_{r}=\frac{1}{n} \sum_{i=1}^{n}\left|f_{r}\left(F\left(x_{i}\right), G\left(F\right), z\left(X_{i}\right)\right)-f_{r}^{*}\left(X_{i}\right)\right|^{2},
\end{aligned}
\end{equation}
where $n$ is the number of sample points, $X_{i} \in \mathbb{R}^{3}$ is a point in the camera coordinate system, $F(x)=g(I(x))$ is the local image feature located at $x$, $G(F)$ is the global image feature of the room background and $F$ is the whole feature map produced by Hourglass network. 
The local and global image features are both from a stacked hourglass network~\cite{newell2016stacked}, but an extra global encoder $G$ is needed to encode the whole feature map $F$ to the global feature. 
The obtained implicit room background can be easily converted to an explicit mesh via marching cube algorithm.

\subsection{Indoor Object Reconstruction} 
As discussed in \cref{sec:instpifu}, due to the heavy occlusions between indoor objects, directly applying PIFu for instance reconstruction suffers from the problem of ambiguous local features.
We adopt the proposed \methodName, which applies \attentionmodule for feature filtering, to reconstruct the indoor objects.
We define the ground-truth surface of an indoor object as the room background (see~\cref{eq:level_set}).
The \methodName $\instalignfunction$ is also trained by minimizing the average of MSE:
\begin{equation}
\begin{aligned}
\mathcal{L}_{o}=\frac{1}{n} \sum_{i=1}^{n}
    \left|\instalignfunction\left(\channelwisefeature\left(x_{i}\right), \globalinstancefeatureabbr , z\left(X_{i}\right)\right) - \instalignfunction^{*}\left(X_{i}\right)\right|^{2},
\end{aligned}
\end{equation}
where $X_{i} \in \mathbb{R}^{3}$ is a point in the canonical coordinate system. 
Note that the projection from $X_{i}$ to $x_{i}$ is different from the original PIFu. Because $X_{i}$ is in object coordinate system, extra camera and object poses are needed when projecting. We follow~\cite{im3d} to predict these parameters for projecting.
The channel-wise attention layer is implemented as MLPs. 
During training, we add an extra instance mask loss for the instance-aligned attention module to enforce the feature to be constrained on the corresponding instance mask. The mask loss is simply implemented by the MSE between the predicted mask and the ground truth.

\subsection{Scene Compositing}
The room background is obtained in the camera coordinate system, while the objects are recovered in their canonical coordinate system to ease the learning of reconstructing indoor objects with various poses and scales. To embed objects into the scene together with the room background, the camera pose $\mathbf{R}(\beta, \gamma)$ and object bounding box parameters $(\delta, d, s, \theta)$ are required. We use similar camera estimator and 3D object detector to predict above parameters as~\cite{total3d,im3d}. Additionally, the Scene Graph Convolutional Network proposed in \cite{im3d} is also used in our work to improve the performance of camera and object pose estimation. Note that we use perspective camera model.

\section{Experiment}
\label{sec:experiment}

\subsection{Experiment Setup}

\paragraph{Datasets} 
We conduct experiments on both synthetic and real datasets. The proposed pipeline is trained on \threeDfront \cite{3d-front} which is a large-scale repository of synthetic indoor scenes, consisting of professionally designed rooms populated by 3D furniture models with high-quality geometry and texture in various styles. 
The furniture models come from \threeDfuture \cite{3d-future}. 
We use about 20K scene images for training and 6K for testing, where more than 16K objects from \threeDfuture are included. 
Following \cite{total3d,im3d}, we also evaluate our method on real-world datasets: \sunrgbd \cite{song2015sun} and \pixthreeD \cite{sun2018pix3d}.

\paragraph{Metrics} 
We adopt the commonly used Chamfer distance (CD) to evaluate the background reconstruction, as it is difficult to compare our background results with layout Intersection over Union (IoU)~\cite{total3d,im3d,yang2022learning,stekovic2020general} (detailed reasons in Sec.~\ref{sec:comp_back}).
The reconstructed indoor objects are evaluated with CD and F-Score~\cite{wang2018pixel2mesh,total3d,im3d}.

\subsection{Evaluation on Room Background Estimation}
\label{sec:comp_back}
We first evaluate the effectiveness of our room background estimation module. 
Layout IoU is a commonly used metric when comparing the room background. It is computed using the layout structure of the whole room. However, our method only reconstructs partial room background within the camera view. Hence, to compare our room
background results with existing methods quantitatively, we firstly sample 10K points within the camera frustum from the reconstructed background in representations of bounding box~\cite{im3d}, depth map~\cite{tulsiani2018factoring,adabins}, plane sets~\cite{liu2019planercnn} and our implicit surface, then compute CD with points on ground truth background. 
We choose to compare with PlaneRCNN~\cite{liu2019planercnn} since it is popular and has decent performance in plane estimation. 
Because Factored3D~\cite{tulsiani2018factoring} is based on depth estimation, we also compare it with Adabins~\cite{adabins} that is the state-of-the-art depth estimation approach.
Quantitative comparisons in~\cref{compare:background} shows the superiority of our method in detailed background recovery. 
Visual results of background reconstruction on \threeDfront and \sunrgbd show that our method can recover the geometry details of the room background (see \cref{fig:scene_recon}). 
More visual comparisons are given in the Supplementary Material.

\begin{table}[htbp]
	\begin{center}
	\small
		\begin{tabular}{|l|c|c|c|c|c|}
			\hline
			Method  & Factored3D~\cite{tulsiani2018factoring}& Adabins~\cite{adabins} & Im3D~\cite{im3d} & PlaneRCNN~\cite{liu2019planercnn} & Ours\\
			\hline
			CD on \threeDfront $\downarrow$  & 0.697 & 0.573 & 1.974 & 0.717  & \textbf{0.481}\\
			\hline
		\end{tabular}
		\caption{Quantitative comparisons of room background estimation on \threeDfront.}
	    \label{compare:background}
	\end{center}
\end{table}

\begin{figure*}[t]
	\centering
	\includegraphics[width=1.0\textwidth]  
	{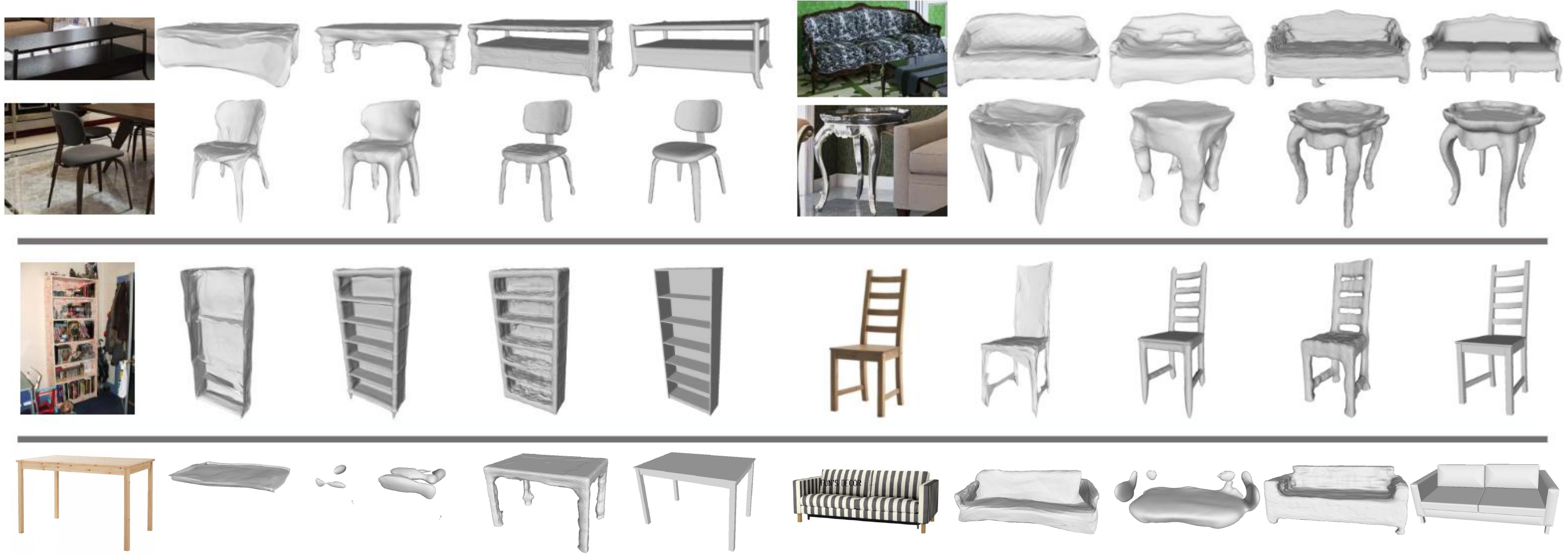}
	\caption{Qualitative comparisons of indoor object reconstruction. From left to right of every quintuplet: (1) Input images and results from (2) MGN \cite{total3d}, (3) LDIF \cite{im3d}, (4) Ours, (5) Ground truth. 
	The first two rows are compared on \threeDfuture, and the last two rows are on \pixthreeD. Note that results of the last row are generated by models trained and tested on non-overlapped split.}
	\label{fig:obj_compare}
\end{figure*}

\subsection{Evaluation on Indoor Object Reconstruction}
We compare our \methodName against the MGN of Total3D~\cite{total3d} and the LIEN of Im3D~\cite{im3d} on indoor object reconstruction.
Quantitative and qualitative comparisons are shown on both \threeDfuture and \pixthreeD. 
Furthermore, we also train and test these object reconstruction networks on \pixthreeD with a non-overlapped split to evaluate their generalization ability.
CD is used to evaluate on the 10K points sampled from the reconstructed mesh after being aligned with the ground-truth using ICP. Note that results generated by InstPIFu and LIEN are in implicit representation which are converted to mesh using marching cube algorithm with a resolution of $256$.

\paragraph{Evaluation on \threeDfuture}
\cref{compare:object_recon_3dfuture} summarizes the quantitative results on \threeDfuture evaluated on $2000$ indoor objects in $8$ different categories. 
We use scene images in \threeDfront as the input for our \methodName, and cropped patches by ground-truth 2D bounding boxes (following~\cite{total3d,im3d}) from every scene image as the input for MGN and LIEN. 
In these input images, object occlusions often occur. And thanks to the use of the instance-aligned feature, our method achieves the best on F-Score and shows decent results on CD (see \cref{compare:object_recon_3dfuture}). 
Although explicit methods like MGN achieve better CD loss as they directly optimize the CD loss during training, the reconstructed meshes lack details \cite{mescheder2019occupancy,Junyi,xu2019disn}. Also, MGN can not generate watertight mesh which is desired in object reconstruction. 
\cref{fig:obj_compare} (first two rows) shows that the results of our method have the most similar appearances to objects in the input images.

\paragraph{Comparison on \pixthreeD}
Quantitative results on \pixthreeD using the train/test split in~\cite{total3d} are shown in \cref{compare:object_recon_pix3d}, where LIEN and MGN achieve better than ours. 
The major reason is that LIEN and MGN tend to be over-fitting on \pixthreeD which has only about 400 shapes. Because the split in~\cite{total3d} is based on different images, and all shapes in testing dataset also occur in training dataset. Also, the usage of pixel-aligned local feature makes our model achieve better generalization ability, but weaken the fitting performance.
Nevertheless, our method still achieves comparable qualitative results (see the third row in \cref{fig:obj_compare}).

\begin{table}[!h]
	\begin{center}
	\scriptsize
		\resizebox{1.0\linewidth}{!}{\begin{tabular}{|l|c| c| c| c| c |c| c| c |c|}
			\hline
			Method & bed & chair & sofa & table & desk & nightstand & cabinet & bookshelf & mean $\downarrow$ / $\uparrow$\\
			\hline
			MGN~\cite{total3d} & \textbf{15.48} / 46.81 & \textbf{11.67} / 57.49 & 8.72 / 64.61 & \textbf{20.90} / 49.80 & \textbf{17.59} / 46.82 & 17.11 / 47.91 & 13.13 / 54.18 & 10.21 / 54.55 & \textbf{14.07} / 55.64\\
			LIEN~\cite{im3d} & 16.81 / 44.28 & 41.40 / 31.61 & 9.51 / 61.40 & 35.65 / 43.22 & 26.63 / 37.04 & 16.78 / 50.76 & 7.44 / 69.21 & 11.70 / 55.33 & 28.52 / 45.63\\
			Ours & 18.17 / \textbf{47.85} & 14.06 / \textbf{59.08} & \textbf{7.66} / \textbf{67.60} & 23.25 / \textbf{56.43} & 33.33 / \textbf{48.49} & \textbf{11.73} / \textbf{57.14} & \textbf{6.04} / \textbf{73.32} & \textbf{8.03} / \textbf{66.13} & 14.46 / \textbf{61.32}\\
			\hline
		\end{tabular}}
		\caption{Quantitative comparisons of object reconstruction on \threeDfuture (CD / F-Score). The values of CD are in units of $10^{-3}$.}
	\label{compare:object_recon_3dfuture}
	\end{center}

	\begin{center}
	\scriptsize
 	\resizebox{1.0\linewidth}{!}{
        \begin{tabular}{|l|c| c| c| c| c |c| c| c |c|c|}
			\hline
			Split in~\cite{total3d} & bed & bookcase & chair & desk & sofa & table & tool & wardrobe & misc & mean $\downarrow$ / $\uparrow$\\
			\hline
			MGN~\cite{total3d} & 5.99 / \textbf{78.08} & 6.56 / 62.98 & \textbf{5.32} / \textbf{72.73} & \textbf{5.93} / 75.04 & \textbf{3.36} / \textbf{79.64} & 14.19 / 65.27 & 3.12 / \textbf{81.17} & 3.83 / 85.51 & 26.93 / 46.76 & 6.84 / \textbf{73.18}\\
			LIEN~\cite{im3d} & \textbf{4.11} / 65.26 & \textbf{3.96} / 46.05 & 5.45 / 59.84 & 7.85 / \textbf{76.03} & 5.61 / 64.02 & \textbf{11.73} / \textbf{72.28} & \textbf{2.39} / 36.09 & 4.31 / 58.59 & \textbf{24.65} / \textbf{57.50} & \textbf{6.72} / 63.96\\
			Ours & 9.52 / 59.47 & 4.38 / \textbf{73.25} & 14.40 / 48.26 &  13.70 / 64.24 & 8.21 / 57.17 & 22.6 / 57.52 & 7.76 / 69.36 & \textbf{3.67} / \textbf{87.36} & 30.32 / 35.05 & 13.60 / 56.07 \\
			\hline
			Non-overlapped Split & bed & bookcase & chair & desk & sofa & table & tool & wardrobe & misc & mean $\downarrow$ / $\uparrow$\\
			\hline
 			MGN~\cite{total3d} & 22.91 / 34.69 & 36.61 / 28.42 & 56.47 / \textbf{35.67} & 33.95 / 34.90 & 9.27 / 51.15 & 81.19 / 17.05 & 94.70 / 57.16 & 10.43 / 52.04 & 137.5 / 10.41 &  44.32 / 36.20\\
			LIEN~\cite{im3d} & 11.88 / 37.13  & 29.61 / 15.51  & 40.01 / 25.70  & 65.36 / 26.01  & 10.54 / 49.71  & 146.13 / 21.16  & 29.63 / 5.85  & 4.88 / 59.46  & 144.06 / 11.04  & 51.31 / 31.45 \\
			Ours & \textbf{10.90} / \textbf{54.99}  & \textbf{7.55} / \textbf{62.26}  & \textbf{32.44} / 35.30  & \textbf{22.09} / \textbf{47.30}  & \textbf{8.13} / \textbf{56.54}  & \textbf{45.82} / \textbf{37.51}  & \textbf{10.29} / \textbf{64.24} & \textbf{1.29} / \textbf{94.62}  & \textbf{47.31} / \textbf{27.03} & \textbf{24.65} / \textbf{45.62}  \\
			\hline
		\end{tabular}}
 		\caption{Quantitative comparisons of object reconstruction on \pixthreeD with split in~\cite{total3d} and non-overlapped split.}
	    \label{compare:object_recon_pix3d}
	\end{center}
\end{table}

\paragraph{Comparison of generalization}
To compare the generalization ability of the above three object reconstruction networks, we re-split \pixthreeD based on different shapes (70\% for training and 30\% for testing), which ensures that all shapes in testing dataset have not been seen when training (non-overlapped split).
Quantitative results are shown in \cref{compare:object_recon_pix3d}, where our method achieves the best result due to the use of local image features. 
In contrast, MGN and LIEN suffer from over-fitting caused by global image features. 
Qualitative results in \cref{fig:obj_compare} (the last row) give the same conclusion, where objects reconstructed by MGN and LIEN are coarse shapes. More results are shown in the supplementary material.

\begin{figure}[htbp]
	\centering
	\includegraphics[width=0.98\textwidth]  
	{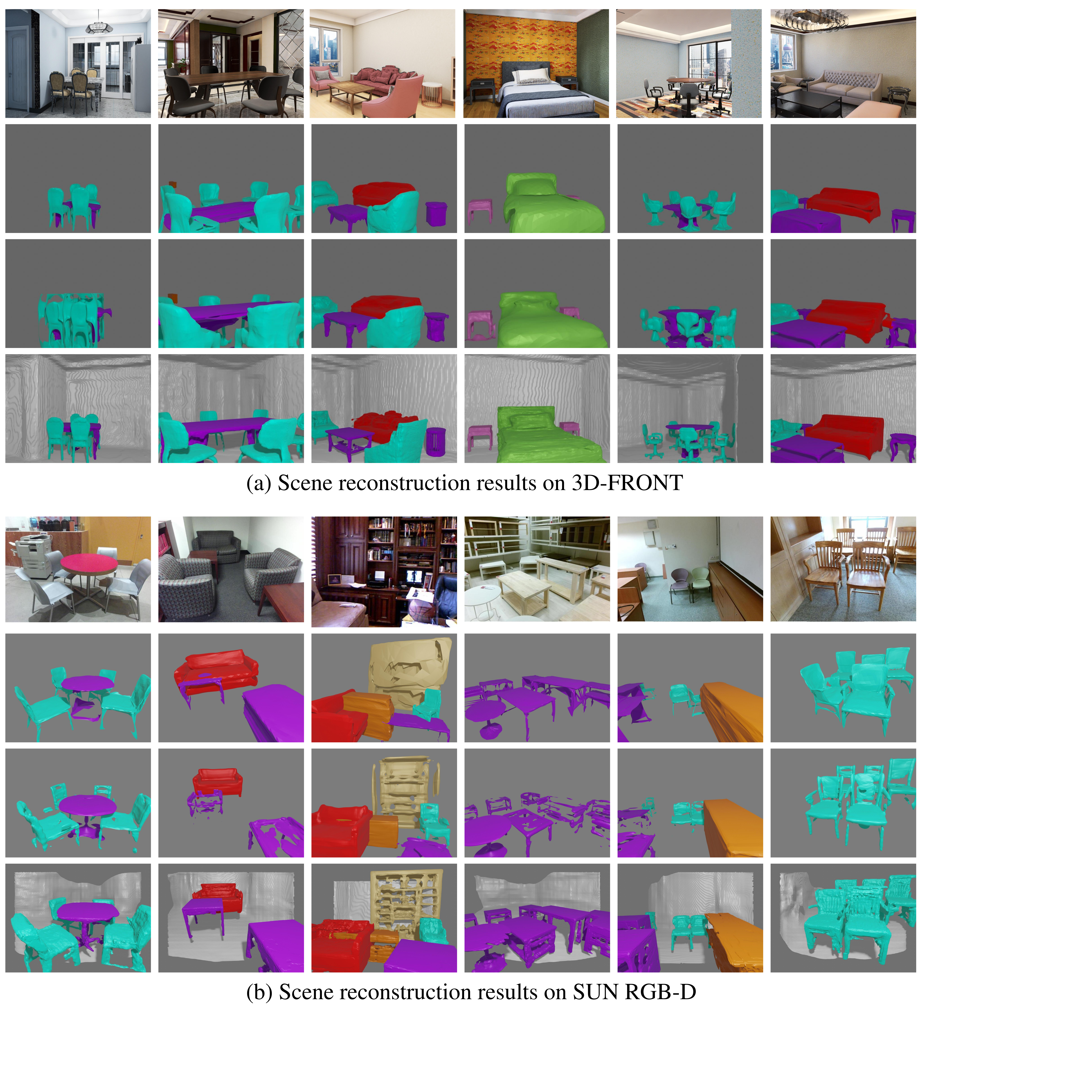}
	\caption{Qualitative comparisons of holistic scene reconstruction. From the first row to the last: the input image, scene reconstruction results of Total3D, Im3D and ours. Note that the first four rows are compared on \threeDfront and the rest are on \sunrgbd.}
	\label{fig:scene_recon}
\end{figure}

\begin{figure}[htbp]
    \centering
	\includegraphics[width=0.98\linewidth]{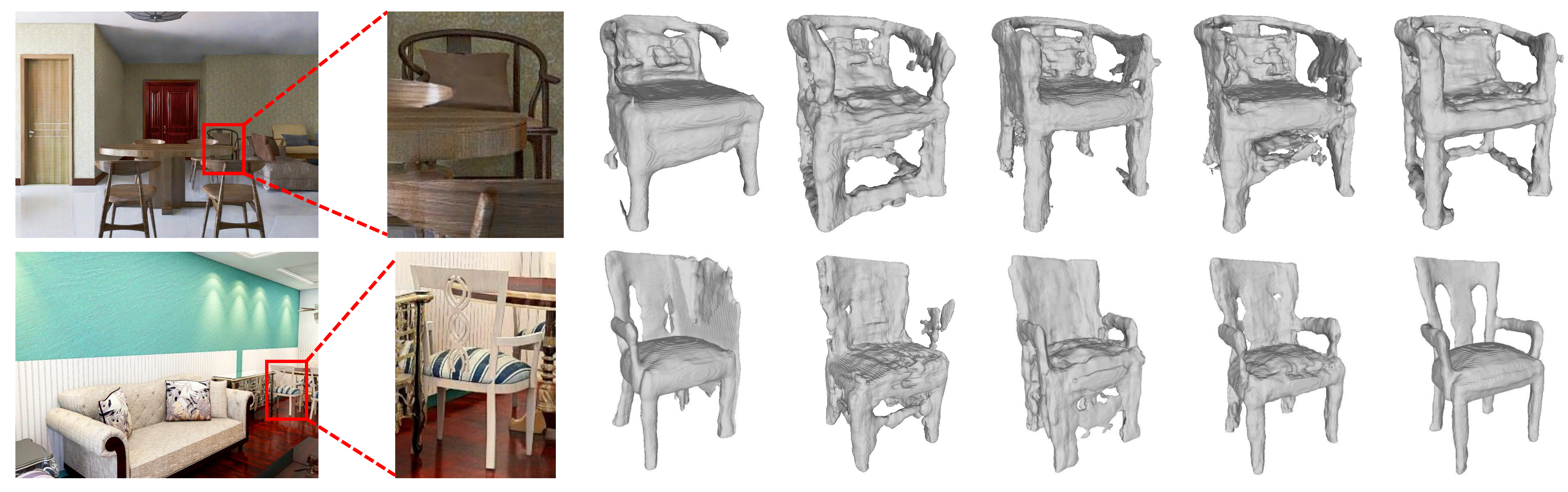}
	\caption{Visual comparisons for ablation study. From left to right: the input image, results of $\bm{Baseline}$, $\bm{C_0}$, $\bm{C_1}$, $\bm{C_2}$ and $\bm{Full}$.}
	\label{fig:ablation}
\end{figure}

\subsection{Qualitative Result of Holistic Scene Reconstruction}
We compare our method with Total3D~\cite{total3d} and Im3D~\cite{im3d} in holistic indoor scene reconstruction on both \threeDfront~\cite{3d-front} and \sunrgbd~\cite{song2015sun} datasets. 
Qualitative comparisons shown in \cref{fig:scene_recon} demonstrate the superiority of our instance-aligned implicit representation.
For fair comparison on \sunrgbd, we first train the \methodName on \threeDfront and \threeDfuture and then finetune it on \pixthreeD. And we also use the predicted 3D object boxes by Im3D. Although our reconstructed scenes on \sunrgbd may have some noisy patches due to the domain gap between the synthetic and the realistic datasets, the results are full of details in both the background and indoor objects, which reveals the good generalization ability of our method to some extend. 

\subsection{Ablation Study}
To better study the effect of instance-aligned implicit representation for indoor object reconstruction, our method is ablated with five configurations:

\begin{itemize}[leftmargin=*]
    \item \textbf{$\bm{Baseline}$:} only pixel-aligned feature is used in object reconstruction.
    \item \textbf{$\bm{C_{0}}$:} pixel-aligned feature + global instance feature.
    \item \textbf{$\bm{C_{1}}$:} $\bm{C_{0}}$ + attentional channel filtering.
    \item \textbf{$\bm{C_{2}}$:} $\bm{C_{0}}$ + spatial-guided supervision.
    \item \textbf{$\bm{Full}$:} $\bm{C_{0}}$ + attentional channel filtering + spatial-guided supervision.
\end{itemize}

\begin{table}[htbp]
	\begin{center}
    	\small
		\begin{tabular}{|l|c|c| c| c| c|}
			\hline
			Method & $\bm{Baseline}$ & $\bm{C_0}$ & $\bm{C_1}$ & $\bm{C_2}$ & $\bm{Full}$\\
			\hline
			CD $\downarrow$ & 17.95& 16.42\textcolor{orange}{(-1.53)} & 15.54\textcolor{orange}{(-2.41)} & 15.28\textcolor{orange}{(-2.67)} & \textbf{14.46}\textcolor{orange}{(-3.49)}\\
			F-Score $\uparrow$ & 56.98& 58.62\textcolor{orange}{(+1.64)} & 60.23\textcolor{orange}{(+3.25)} & 60.56\textcolor{orange}{(+3.58)} & \textbf{61.32}\textcolor{orange}{(+4.34)}\\
			\hline
		\end{tabular}
		\caption{Ablation study for the network architecture.}
    	\label{compare:ablation}
	\end{center}
\end{table}

As the quantitative comparisons shown in \cref{compare:ablation}, our $\bm{Full}$ model achieves the best results on metrics CD and F-Score, where we add the channel-wise attention together with the mask supervision to $\bm{C_0}$. If we remove the anyone of these two modules from $\bm{Full}$, that are $\bm{C_1}$ and $\bm{C_2}$, CD and F-Score both become worse. But $\bm{C_1}$ and $\bm{C_2}$ still perform better than $\bm{C_0}$. This gives us the insight that both of channel-level filtering and spatial guidance help to decouple the feature ambiguity towards occluded objects. And the comparisons between the $\bm{Baseline}$ and $\bm{C_0}$ show that concatenating global instance feature with pixel-aligned local feature is helpful for indoor object reconstruction. But from the comparisons of the whole table, we can see that only using global feature to tackle the ambiguity in occlusion region is not enough. Same conclusions can be drawn by the visual comparisons in \cref{fig:ablation}.

\section{Conclusion}
We have introduced a new method based on implicit representation, called \methodName, for holistic and detailed 3D indoor scene reconstruction from a single image. 
To resolve the problem of ambiguous local features caused by object occlusions in an indoor scene, we proposed an instance-aligned attention module to effectively disentangle the mixed features for accurate instance shape reconstruction.
Moreover, our method is the first to estimate the detailed room background via implicit representation, resulting in a more complete scene reconstruction.
Extensive experiments on both synthetic and real datasets show that our method achieves state-of-the-art results for this problem. 

Although our instance-aligned implicit function enables a more detailed and accurate indoor object reconstruction, the use of local feature makes the joint training of the 3D detection network and object reconstruction network not easy. 
Besides, real-world indoor scene datasets with high-quality 3D ground truth are scarce, and methods trained or finetuned with limited real data perform less well on real-world scenes compared with results on the synthetic scene (see \cref{fig:scene_recon}). 
It would be interesting to explore how to take advantage of the existing large-scale and photo-realistic synthetic datasets for improving the generalization ability of the method.

\paragraph{\textbf{Acknowledgement.}}
The work was supported in part by the National Key R$\&$D Program of China with grant No. 2018YFB1800800, the Basic Research Project No. HZQB-KCZYZ-2021067 of Hetao Shenzhen-HK S$\&$T Cooperation Zone, by Shenzhen Outstanding Talents Training Fund 202002, by Guangdong Research Projects No. 2017ZT07X152 and No. 2019CX01X104, and by the Guangdong Provincial Key Laboratory of Future Networks of Intelligence (Grant No. 2022B12 12010001). It was also supported by NSFC-62172348, NSFC-61902334 and Shenzhen General Project (No. JCYJ20190814112007258). Thanks to the ITSO in CUHKSZ for their High-Performance Computing Services.

\clearpage
\bibliographystyle{splncs04}
\bibliography{egbib}

\clearpage

\appendix 
\counterwithin{figure}{section}
\counterwithin{table}{section}

\centerline{\textbf{\LARGE{-- {Supplementary Material} --}}}
\section{Implementation Details}

\paragraph{Room background estimation} For background estimation, we use ResNet-18 to encode the global room feature, where the input scene image is resized to $256 \times 256$ and the output feature vector has a dimension of $256$. 
As for local features, we adopt the same modified stacked hourglass network as PIFu~\cite{saito2019pifu}, and the size of the input scene image is $484\times648$. 
We also use the same decoder as \cite{saito2019pifu} to be our room  reconstruction PIFu, but change the input dimension of the first layer. 
Different from \cite{saito2019pifu}, we use a positional encoding block like \cite{yu2021pixelnerf} with a frequency of $4$, which makes the occupancy decoding achieve better performance. 

\paragraph{Object reconstruction} For object reconstruction, we first crop the target object using its 2D bounding box and resize it to $256 \times 256$ before feeding it into a ResNet-18 encoder to produce the object global feature. 
We also use this 2D bounding box to perform RoI Alignment \cite{he2017mask} on the feature map of size $256\times64\times64$ encoded by the hourglass network. 
Network architectures for channel-wise attention and mask prediction are all MLPs with $3$ and $4$ layers, respectively. The architecture of the PIFu decoder is the same as the one for room background estimation. 
Attentional Channel Filtering module is constructed by two layers of convolution, followed by a global average pooling to generate a feature vector. Two more layers of MLP and a Sigmoid activation is applied to generate the channel-wise attentional weight. The mask segmentation module has four layers of convolution with kernel size 1, all layers is followed with ReLU activation except the last layer that uses Sigmoid activation. The whole object reconstruction pipeline is trained jointly.
As our InstPIFu is trained with cross-category data, the shape decoder needs an extra category code like the MGN in Total3D \cite{total3d}. 
The loss weight ratio of the occupancy loss to the mask loss is $1:1$ during training.

\paragraph{More Implementation Details} 
We use a similar camera estimator, 2D and 3D detection networks as Total3D~\cite{total3d}. 
The global image encoders $G$ and $G^{'}$ in Fig. 2 are both simple MLPs after a few pooling and convolution operations. 
For the local feature extractor and the PIFu decoders, we adopt the same structures as in \cite{saito2019pifu}. 
Our InstPIFu is trained on \threeDfront, where $W_r = H_r = 64$ and $L_c = 256$. 
3D points used to train InstPIFu are sampled around the mesh surface randomly as \cite{saito2019pifu} and within the bounding box uniformly with a ratio of $1:1$. 
Training is conducted using the batch size of $16$ for $100$ epochs on two NVIDIA RTX3090Ti using $80$ hours.
The learning rate is initialized as $0.0001$, and decayed by a factor of 0.2 in the $50$\textsuperscript{th} and $80$\textsuperscript{th} epochs.

\section{More Quantitative Comparisons}
\paragraph{Coordinate system}
As we mention in Section 4.3, indoor shapes are recovered in canonical coordinate to ease the learning of reconstructing indoor objects with various poses and scales. \cref{tab:coord} gives the results of object reconstruction in camera coordinate, which is worse, justifying the design of our method.

\begin{table}[htpb]
	\begin{center}
	\scriptsize
		\resizebox{1.0\linewidth}{!}{\begin{tabular}{|l|c| c| c| c| c |c| c| c |c|}
			\hline
			Category & bed & chair & sofa & table & desk & nightstand & cabinet & bookshelf & mean $\downarrow$ / $\uparrow$\\
			\hline\hline
			$Ours_{cam}$ & 35.40 / 31.22 & 48.98 / 27.07 & 14.12 / 50.46 & 50.51 / 37.17 & 58.98 / 30.02 & 71.63 / 22.29 & 35.02 / 34.23 & 24.64 / 39.48 & 41.64 / 34.93\\
			$Ours_{cano}$ & \textbf{18.17} / \textbf{47.85} & \textbf{14.06} / \textbf{59.08} & \textbf{7.66} / \textbf{67.60} & \textbf{23.25} / \textbf{56.43} & \textbf{33.33} / \textbf{48.49} & \textbf{11.73} / \textbf{57.14} & \textbf{6.04} / \textbf{73.32} & \textbf{8.03} / \textbf{66.13} & \textbf{14.46} / \textbf{61.32}\\
			\hline
		\end{tabular}}
		\caption{Quantitative comparisons of using different coordinate systems for object reconstruction on \threeDfuture (CD / F-Score). The values of CD are in units of $10^{-3}$.}
	\label{tab:coord}
	\end{center}
\end{table}

\paragraph{Effect of 2D detection accuracy}
We follow Total3D and Im3D to use GT 2D detection during training. There are misaligned bounding boxes when using off-shelf 2D detection while testing. 
\cref{tab:detc} gives the reconstruction results with GT 2D detection and predicted 2D detection by Faster R-CNN trained on MS COCO dataset, and using predicted 2D bounding box causes minor performance drop.

\begin{table}[htpb]
	\begin{center}
	\scriptsize
		\resizebox{1.0\linewidth}{!}{\begin{tabular}{|l|c| c| c| c| c |c| c| c |c|}
			\hline
			Category & bed & chair & sofa & table & desk & nightstand & cabinet & bookshelf & mean $\downarrow$ / $\uparrow$\\
			\hline\hline
			$Ours_{pred}$ & 19.60 / \textbf{47.99} & 18.30 / 55.48 & 8.14 / 64.91 & 25.81 / 50.23 & 56.10 / 41.87 & \textbf{10.16} / 46.70 & 8.59 / 64.35 & \textbf{7.15} / \textbf{69.39} & 15.35 / 58.21\\
			$Ours_{gt}$ & \textbf{18.17} / 47.85 & \textbf{14.06} / \textbf{59.08} & \textbf{7.66} / \textbf{67.60} & \textbf{23.25} / \textbf{56.43} & \textbf{33.33} / \textbf{48.49} & 11.73 / \textbf{57.14} & \textbf{6.04} / \textbf{73.32} & 8.03 / 66.13 & \textbf{14.46} / \textbf{61.32}\\
			\hline
		\end{tabular}}
		\caption{Quantitative comparisons of using different 2D object proposals for the testing of object reconstruction network on \threeDfuture (CD / F-Score).}
	\label{tab:detc}
	\end{center}
\end{table}

\paragraph{Evaluation on the Whole Scene}
We evaluate the reconstruction results of the whole scene including the background and instances of the foreground by CD on 500 images of test scenes of 3D-FRONT, where our approach gets 0.172 and the baseline gets  0.290.

\section{More Qualitative Comparisons}

\begin{figure}[htbp]
    \centering
	\includegraphics[width=0.99\linewidth]{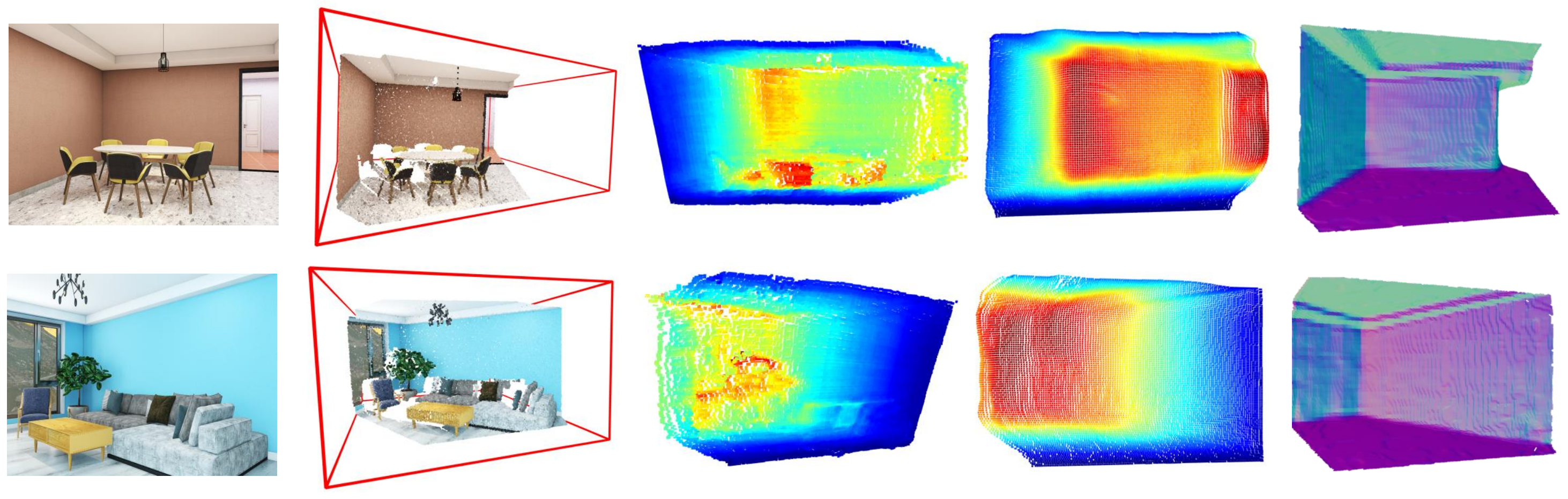}
	\caption{Visual comparisons for background representation, from left to right: scene images, reconstructed 3D room bounding boxes~\cite{im3d}, depth estimation for Factored3D~\cite{tulsiani2018factoring} and Adabins~\cite{adabins}, as well as our results.}
	\label{fig:bg1}
\end{figure}

\begin{figure}[htbp]
    \centering
	\includegraphics[width=0.99\linewidth]{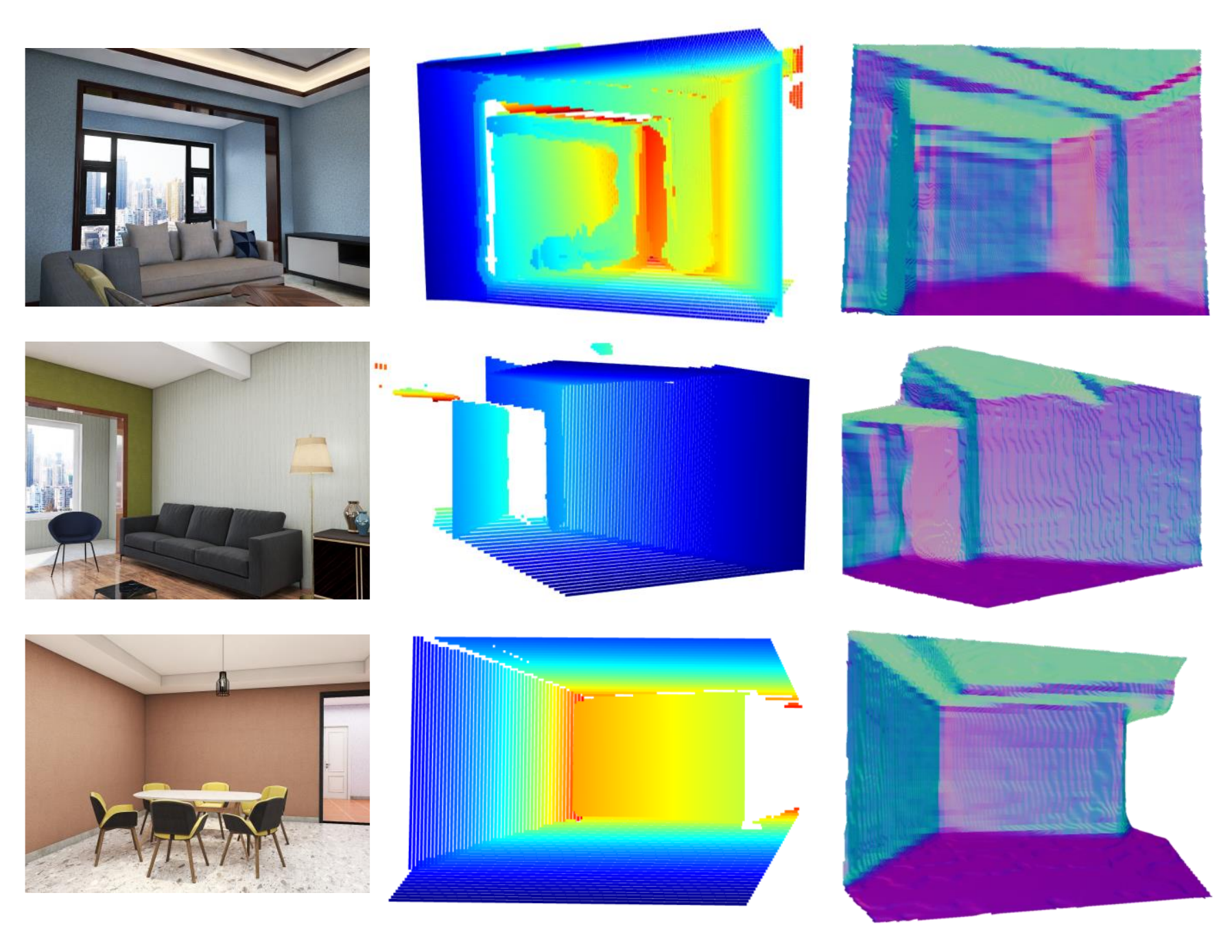}
	\caption{Visual comparisons for background representation between PlaneRCNN~\cite{liu2019planercnn} (middle) and ours (right).}
	\label{fig:bg2}
\end{figure}

\paragraph{Background on 3D-FRONT} More visual results of background on 3D-FRONT~\cite{3d-front} is shown in ~\cref{fig:bg1} and ~\cref{fig:bg2}.

\begin{figure*}[!ht]
	\centering
	\includegraphics[width=0.93\textwidth]  
	{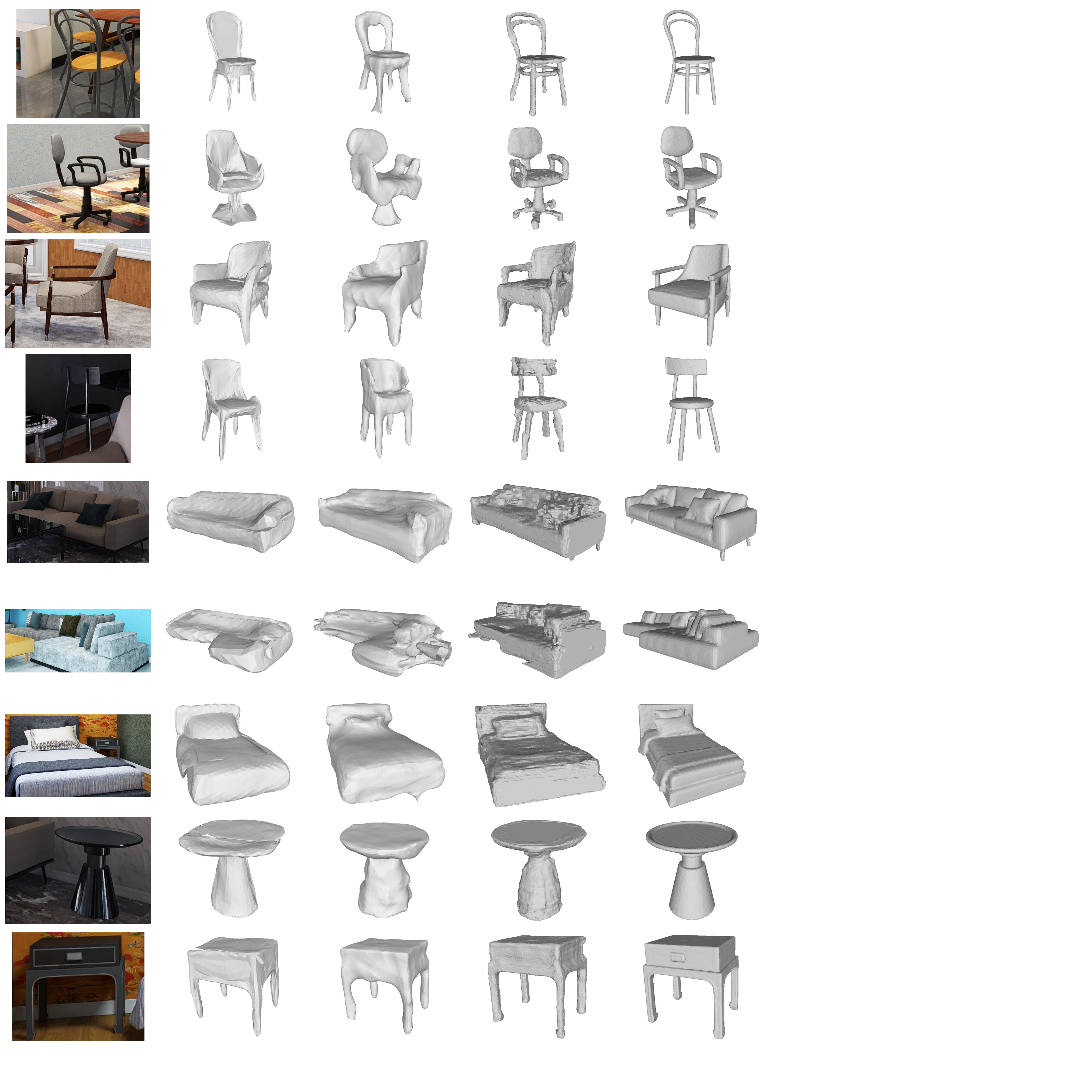}
	\caption{More qualitative comparisons of indoor object reconstruction on 3D-FUTURE. From left to right of every quintuplet: (1) Input image, results from (2) MGN \cite{total3d}, (3) LDIF \cite{im3d}, (4) Ours, and (5) Ground truth.}
	\label{fig:more_3dfuture}
\end{figure*}

\paragraph{3D-FUTURE} More qualitative comparisons of indoor object reconstruction on 3D-FUTURE \cite{3d-future} are shown in \cref{fig:more_3dfuture}.

\begin{figure*}[!ht]
	\centering
	\includegraphics[width=0.99\textwidth]  
	{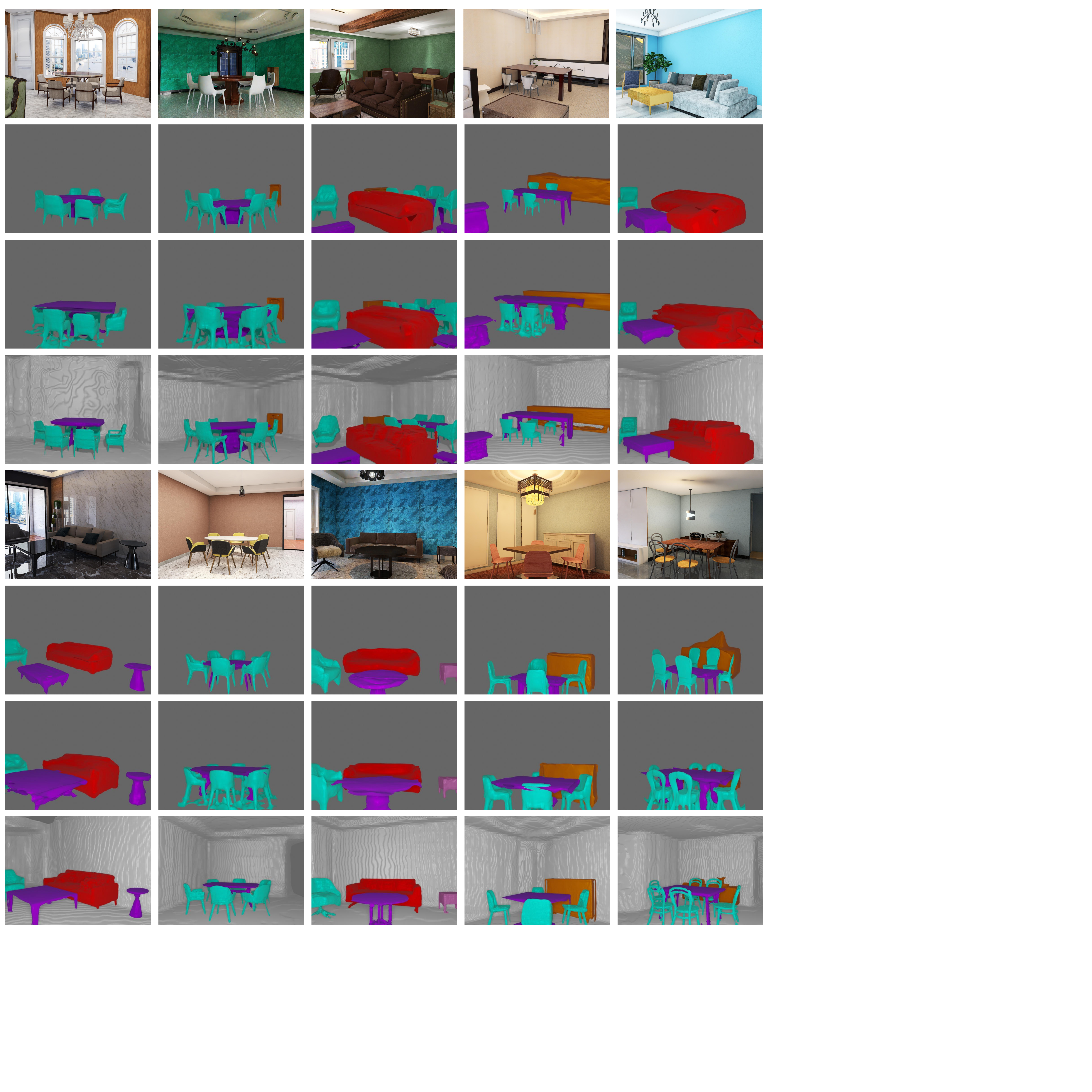}
	\caption{More qualitative comparisons of holistic scene reconstruction on 3D-FRONT. From the first row to the last: the input image, scene reconstruction results of Total3D \cite{total3d}, Im3D \cite{im3d}, and Ours.}
	\label{fig:more_3dfront}
\end{figure*}

\paragraph{3D-FRONT} We present more scene reconstruction results on the testing set of 3D-FRONT \cite{3d-front} in \cref{fig:more_3dfront}.

\begin{figure*}[!ht]
	\centering
	\includegraphics[width=0.99\textwidth]  
	{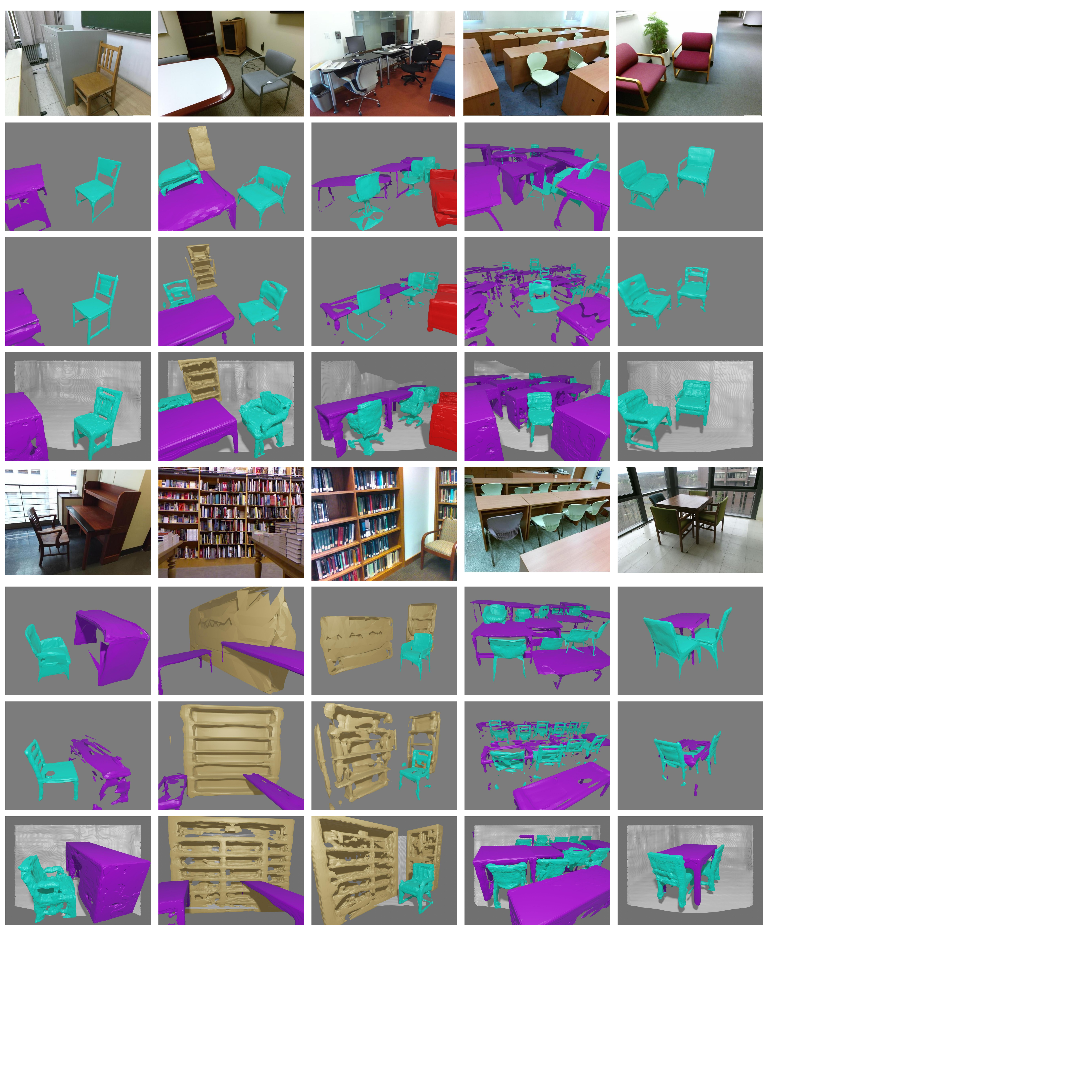}
	\caption{More qualitative comparisons of holistic scene reconstruction on SUN RGB-D. From the first row to the last: the input image, scene reconstruction results of Total3D \cite{total3d}, Im3D \cite{im3d}, and Ours.}
	\label{fig:more_sunrgbd}
\end{figure*}

\paragraph{SUN RGB-D} We present more scene reconstruction results on the testing set of SUN RGB-D \cite{song2015sun} in \cref{fig:more_sunrgbd}.


\end{document}